\documentclass[doc]{apa6}

\usepackage{listings}
\usepackage{float}

\usepackage{amsmath,amssymb}
\usepackage{mathtools}

\usepackage{changepage}

\usepackage[utf8x]{inputenc}

\usepackage{textcomp,marvosym}

\usepackage[numbers, super, sort]{natbib}

\usepackage[table]{xcolor}

\usepackage{array}

\usepackage{lipsum}  

\usepackage{subcaption}

\usepackage{algorithm}
\usepackage[noend]{algpseudocode}
\makeatletter
\def\BState{\State\hskip-\ALG@thistlm}
\makeatother

\title{Meta-Learning to Explore via Memory Density Feedback}
\shorttitle{Exploration via Memory Density Feedback}
\author{Kevin McKee, Eric Alt, Andrew Grebenisan, Mick van Gelderen, Gary Miguel}
\affiliation{Astera Institute}

\abstract{
Exploration algorithms for reinforcement learning typically replace or augment the reward function with an additional ``intrinsic'' reward that trains the agent to seek previously unseen states of the environment.
Here, we consider an exploration algorithm that exploits meta-learning, or learning to learn, such that the agent learns to maximize its exploration progress within a single episode, even between epochs of training.
The agent learns a policy that aims to minimize the probability density of new observations with respect to all of its memories. 
In addition, it receives as feedback evaluations of the current observation density and retains that feedback in a recurrent network.
By remembering trajectories of density, the agent learns to navigate a complex and growing landscape of familiarity in real-time, allowing it to maximize its exploration progress even in completely novel states of the environment for which its policy has not been trained. 
}

\begin{document}
\maketitle

\section*{Introduction}
%WHAT IS EXPLORATION IN RL?
In reinforcement learning (RL), exploration refers to algorithms that induce an agent to observe as much of a given task as possible.
All RL algorithms include some form of random exploration, such as the epsilon-greedy policy or by additionally training to maximize the policy's entropy.
These algorithms are necessary for the agent to find rewarding states and expand its policy, but often fall short when rewards are sparsely distributed, that is, requiring non-obvious and improbable sequences of action.
To improve the generality of RL agents in sparse reward environments, additional exploration algorithms, often called ``curiosity'' are added to the agent, usually in the form of sophisticated objective functions and neural network modules.

Some of such approaches give the agent a distinct enough intrinsic reward that it will systematically learn about the environment and develop general skills in the absence of any extrinsic rewards.
Given that RL environments usually represent only a small, contrived subset of programs for which we might desire autonomous control, it is necessary to develop robust exploration algorithms if we wish to develop artificial intelligence that does not require careful, manual engineering of reward functions for every possible task.

Exploration is likely to be particularly important, even central, to machine learning based approaches to artificial general intelligence that use offline memory replay-based training.
By continually searching for new data irrespective of particular reward functions, an agent collects the prerequisites to maximize any subsequent reward function. 
That may be either through a cumulative “meta-learning” policy such as goal-conditioning, or by reward-specific offline fine-tuning. 
In this study, we develop a method of exploration that maximizes coverage of an environment's observation space while leaving the question of how that data are used to further research.

%WHAT KINDS OF CURIOSITY ARE THERE?
\paragraph{Kinds of Curiosity}
The most popular exploration algorithms each tend to fall into one of three categories by the kind of objective term used: prediction error, reconstruction error, and memory density.
Each of these algorithms works because the target is non-stationary.
Model-based methods promote attainment of un-modeled data, which then used to improve the model.\cite{schmidhuber1991possibility, sekar2020planning}
Model-free methods promote attainment of data that are not yet in memory, but cease to be novel upon collection. 
All methods result in a cyclical progression of novel states through the environment. 

Specifically, prediction error methods use a world model to generate predictions of the environment given possible actions of the agent, then compute the difference in the prediction and the actual result after the action is taken.
While the world model is trained to minimize prediction error, the policy is trained to maximize prediction error, resulting in an agent that pursues environment states and dynamics that are not yet accurately modeled.
This approach pushes the agent to develop a diverse model of the environment and potentially filtering out superficial kinds of novelty.
However, the agent may get stuck observing intrinsically noisy attributes of the environment if the noise variance is large enough, a phenomenon sometimes called the ``noisy TV problem.''

Reconstruction error methods instead train a model to autoencode the environment states.
That is, they encode a compressed representation of the data then reconstruct the data from that compressed representation.
The compression is typically much lower dimensional than the input data, producing an information bottleneck that forces the model to learn any underlying simple structure.
Another approach to curiosity then is to train the policy to maximize observations that cannot be well reconstructed from that simple structure.
This approach does not suffer from environmental noise because the noise does not introduce additional error. 
Rather, samples from the noise distribution are reconstructed, whatever they may be, which is likely to be an exhaustible process.
Although formulated in a slightly different way, Random Network Distillation (RND)\cite{burda2018exploration} falls into this category, as only concurrent observations are modeled.

If no model is present, the agent may use a memory buffer of previously seen observations to determine novelty.
The most prominent example of this may be Go-Explore \cite{ecoffet2019go, ecoffet2021first}, and offshoots such Latent Go-Explore \cite{gallouedec2023cell}.
These methods require some calculation of novelty.
For discrete environments, it may be as simple as counting the number of times each state has been seen.
For continuous environments, it is necessary to apply a density estimator.
These methods have the advantage of applicability in simpler agents without the need to exhaustively train a world model.
They are also unlikely to suffer from environmental noise, as the density of memories will increase with observation of the noise distribution and eventually be less rewarding than new states altogether. 

%WHAT ENVIRONMENTAL FACTORS INFLUENCE EXPLORATION?
\paragraph{Environmental Attributes}
There are several environmental factors that categorically determine the effectiveness of each exploration algorithm.
Because algorithms differ in their effectiveness with respect to these factors, they should be treated as complementary forms of curiosity.

The above mentioned ``noisy TV problem'' is one challenge.
Interesting environments are likely to have stochastic elements, and so prediction error alone may be inadequate on its own.
In more extreme cases, major features of the environment may be randomized per episode, such as layouts, obstacles, and reward conditions.
The result is that the optimal path to any given reward or state of the environment is unpredictable.
In such highly randomized tasks, memory density methods may result in superficial exploration and simplistic policies because little is needed to produce observations that are novel at face value.

An environment's observations may not be complex and varied, perhaps because the novelty is not in the particular presentation of the environment, but in the rules or dynamics governing what is presented.
In that case, reconstruction error may become ineffective very quickly, having little for the autoencoder to learn.

Finally, the environment may have an episodic structure or not; if episodic, then the explorer must learn to return to its frontier and extend its progress. 
If not, it must learn to return old paths not taken whenever possible.
This requires potentially sophisticated memory systems and navigation according to both reliable environmental landmarks and internally generated novelty feedback. 

%PROPOSAL: meta-learning OF EXPLORATION
\paragraph{Meta-Learning to Explore}
In this study, we extend the idea of curiosity by allowing the agent to meta-learn the maximization of novelty.
To do this, we follow a simple concept and implementation of meta-learning: the agent takes its own actions and reward as feedback, maps them into short-term memory, and learns a policy with respect to trajectories of feedback.\cite{wang2016learning}
The goal is to produce an agent that responds to trajectories of novelty calculated in real-time, such that the agent continues to optimally explore even after it is outside of its stored distribution of environment states.
To discuss and implement this, we consider a purely observation-conditioned policy (OCP), a purely feedback-conditioned policy (FCP), and the combination of both.
The resulting agent should explore efficiently in both fixed and randomized environments because it is able to use as reference both static, reliable states and feedback amidst unreliable random states to navigate.
For both simplicity, generality, and robustness to noise, we use a model-free, memory density based approach, though in principle, meta-learning may be applied to model-based methods as well.

We hypothesize that if novelty is provided as feedback along with actions, the agent will learn to explore more efficiently in general with continued training, leading to acceleration in task coverage.
Second, if observations are provided along with feedback (both FCP and OCP), the agent will leverage both general, reactive methods of exploring and exploit regularities in the observation space, improving performance above either OCP or FCP alone on all tasks.

\section*{Tasks}
To make specific comparisons of this algorithm with previous results, all diagnostic environments were variants of the continuous maze presented in Latent Go-Explore paper.\cite{gallouedec2023cell}
In keeping with the Latent Go-Explore comparisons, we also use the continuous version of the agent with DDPG (See Appendix A).
This maze functions as an analogy for environments generally, requiring lengthy chains of sub-goal locations to reach the furthest point.

\paragraph{Fixed maze}
The first test is taken directly from Latent Go-Explore.\cite{ecoffet2019go}
The agent receives only its own coordinates $(x,y)\in [-12, 12]$ and internally generated feedback as input.
Its goal is to maximize coverage of the maze via intrinsic rewards, without having access to a coverage metric.
There are no other rewards for the task.
See results from the original paper for the performance of random noise and several other exploration algorithms on this task. 

\paragraph{Random maze}
We used Prim's algorithm\cite{prim1957shortest} to generate a new random maze every episode.
As there are no reliable paths to any particular location, the agent cannot benefit from exploration methods that rely primarily on repeatable observation states, such as with Go-Explore.

\paragraph{Continual maze} 
We generate a larger, more complex maze, and removing episodic resets to the center.
Instead, the agent explores continually.
Without restarts, the agent is challenged to backtrack and re-explore past areas.

\paragraph{Noisy maze} 
An additional input element is added to the task.
The element contains random noise when the agent enters one of four zones in the maze, and is zero otherwise.
This challenges the algorithm to deal with the ''noisy TV problem,'' in which prediction-based exploration algorithms lead the agent to fixate on intrinsically unpredictable inputs.

\paragraph{Non-Euclidean Graph Mazes} 
A potential weakness of this algorithm concerns its use of Euclidean distance, which has diminishing ability to distinguish nearest from furthest neighbors in high dimensional space \cite{aggarwal2001surprising}.
To determine robustness of the algorithm to high-dimensional task spaces not organized in a Euclidian grid, we implemented a generator of discrete, non-Euclidean graph mazes.
The maze generator used Wilson's algorithm\cite{wilson1996generating} to produce transition tables connecting state-action pairs to subsequent states.
Our exploration algorithm was tested on random mazes with four actions and observation dimensions of 2, 8, and 32.
The total sizes of the maze projected to 2-dimensions was 25 $\times$ 25, giving them four as many cells as the fixed 2D continuous maze.

\section*{Model}
The proposed explorer model can be implemented for discrete actions with DQN\cite{mnih2013playing} or for continuous actions with DDPG.\cite{lillicrap2015continuous}
We present the algorithm here in terms of DQN for simplicity but perform our base experiments with the continuous DDPG-based implementation.
The continuous DDPG model is given in Appendix A.
The model hyperparameters used in our experiments are given in Appendix B.

The proposed design follows upon our previous work investigating choices of recurrent neural network (RNN) for best-performing short-term memory in meta-learning contexts, along with other previous work with similar findings. \cite{subramoney2021reservoirs, mckee2024reservoir}
Observations and feedback are fed to a reservoir, producing a compression of their running history.\cite{lukovsevivcius2009reservoir, jaeger2004harnessing} 
The reservoir state maps to the policy via a fully connected multi-layer network.
In environments with 1D observation spaces, only the weights of the output network are trained. 
The simplest form of reservoir computer to implement given existing tools is the Echo State Network (ESN), which is a recurrent network with $\tanh$ as the activation function.\cite{jaeger2001short, jaeger2004harnessing, jaeger2001echo}
For 2D observation spaces, a convolutional network or other choice suitable to the task can be added upstream from the reservoir to preprocess the observations.

\begin{figure}[!ht]
    \centering
    \includegraphics[width=.75\linewidth]{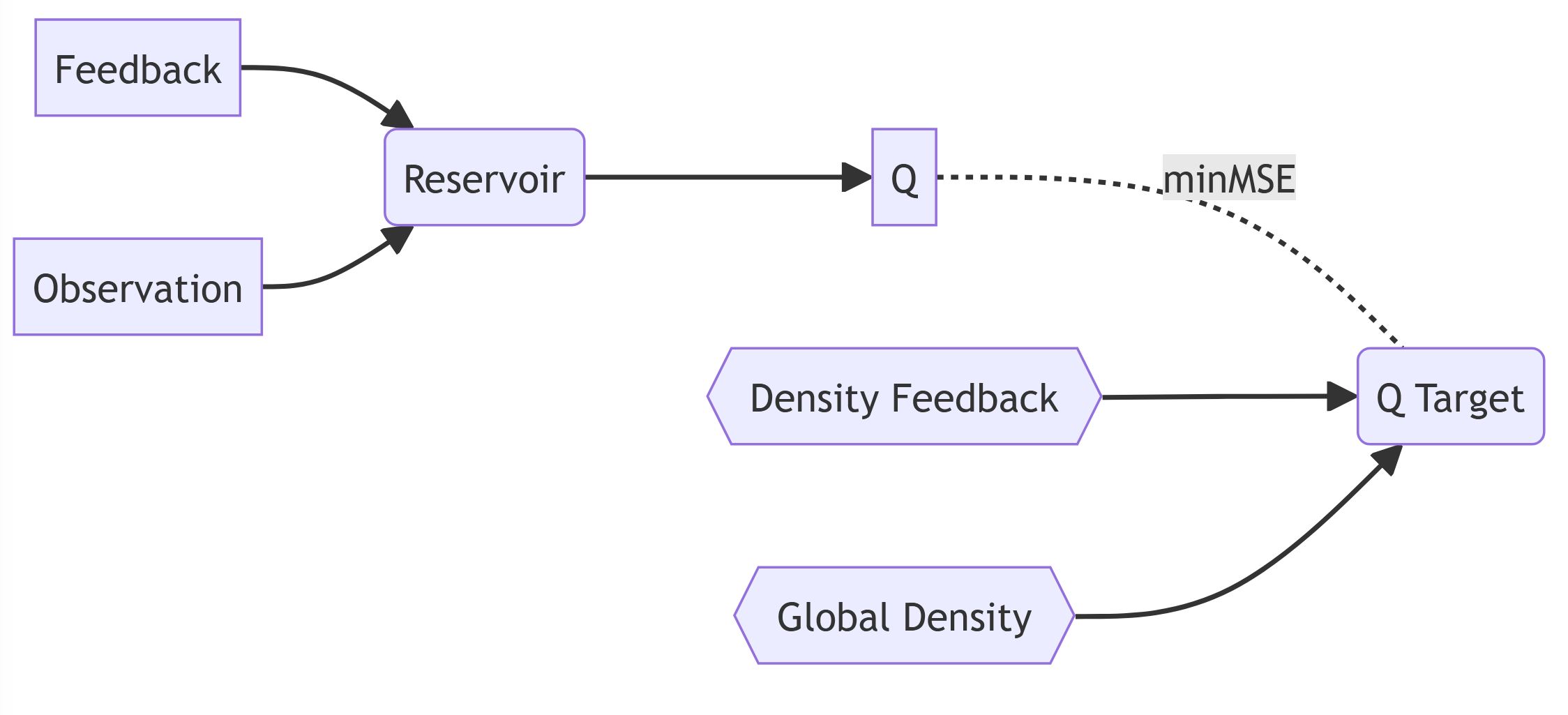}
    \caption{DQN based explorer for discrete actions}
    \label{fig:model}
\end{figure}

The feedback passed to the reservoir includes the previous action, any previous task reward, and the negative density of the current observation normalized to the unit interval.
These inputs allow the agent to learn overall exploration heuristics regardless of the observations, in case the observations are highly random or uninformative to the maximization of novelty.

Because the model trains on a cumulative memory buffer, let $t$ denote total time over all episodes, not just within episode.
Let $X$ refer to the buffer containing all stored observations $x$ and $D$ refer to the buffer containing all observation densities $d$ computed online.
The discrete model is just
\begin{align}
z_t &= \text{ESN}([a_{t-1}, r_{t}, \bar d_t, x_t]), \\
q_t &= \text{MLP}(z_t), \\
a_t &= \arg \max q_t,
\end{align}
where $\bar d_t = d_t/\max(D_t)$.
MLP refers to a standard multi-layer perceptron, and ESN refers to an Echo State Network, which is a simple RNN with fixed, random, sparsified weights that are normalized to have a spectral density close to 1.0.\cite{jaeger2001short}

To train the model, we draw one or more complete episodes from the memory buffer and from it compute the target
\begin{align}
q_{t+1}' &= r_{t+1} + \beta_d d_{t+1} + \beta_g g_{t+1} + \gamma \max q_t(z_{t+1}'),
\end{align}
and minimize the mean squared error objective $\mathbb{E}[||q_{a}-q'||^2]$. 
Online negative density $d_t$ and offline negative density $g_t$ are calculated using the k-nearest neighbor method used by Latent Go-Explore:\cite{kung2012optimal, gallouedec2023cell}
\begin{align}
\mathcal{D}(x, Y, k) &= \text{sort}(||x-y_i||^2\quad \forall y_i \in Y )_k.
\end{align}
We take the Euclidian distance of each observation to its $k^\text{th}$ nearest neighbor in the memory buffer, where $k$ is a hyperparameter.
Higher values of $k$ result in a smoother density function.
Specifically, we compute the Euclidian distance of an input vector $x$ to each member in the set of vectors $Y$,
sort the resulting distances, and take the $k^\text{th}$ member.
This Euclidian distance is monotonic with the negative probability density.\cite{gallouedec2023cell}

As the agent interacts with the environment, observation $x_t$ is appended to the memory buffer $X_{t-1}$ to get $X_t$ such that $x_0 \hdots x_t \in X_t$.
An online density calculation is performed to obtain $d_t = \mathcal{D}(x_t, X_{t-1}, k)$.
$d_t$ is appended to memory buffer $D_{t-1}$ to get $D_t$.
Hence, at offline training time, we have $d_t \in D_t$, the densities computed with respect to memory up to time $t$, and we have $g_t \in \mathcal{D}(X, X, k)$, which is an up-to-date calculation of all densities with respect to the complete set of memories.

\begin{algorithm}[!ht]
\caption{Recurrent DQN Explorer}\label{algorithm}
\begin{algorithmic}[1]
\Procedure{}{}

\State \text{Initialize buffers} $X, R, D, A$ 
\State \text{Initialize recurrent policy} $Q(x, \theta)$ 
\State \text{Initialize target policy} $Q'(x, \theta')$ 
\State $a_0 \gets 0$

\BState \emph{loop}
\State $x_0, r_{0}, done \gets \text{Initialize environment } \psi$

\While{not $done$}:
\State $d_t \gets \mathcal{D}(x_t, X_{t-1},k) \quad\quad\text{Negative density calculation}$
\State $\bar d_t \gets d_t / \max(D_{t-1})$
\State $q_t \gets Q([x_t, r_{t}, \bar d_{t}], \theta) \quad\quad\text{Model step}$
\State $a_t \gets \arg \max q_t$
\State $x_{t+1}, r_{t+1}, done \gets \psi(a_t) \quad\quad\text{Environment step}$
\State $X_t \gets [X_{t-1}, x_t] \quad\quad\text{Observation buffer}$
\State $R_t \gets [R_{t-1}, r_t] \quad\quad\text{Reward buffer}$
\State $D_t \gets [D_{t-1}, d_t] \quad\quad\text{Density buffer}$
\State $A_t \gets [A_{t-1}, a_t] \quad\quad\text{Action buffer}$
\State $t \gets t + 1$
\EndWhile

\State $G \gets \mathcal{D}(X, X, k) \quad\quad\text{Offline goal rewards}$ 
\For{Epochs}{}
\For{Training steps per epoch}{}
% \State $\text{Sample episodes} (x\in X^s \subset X, d \in D^s \subset D, r\in R^s \subset R, a \in A^s \subset A, g \in G^s \subset G, )$
\State $\text{Sample episodes} \quad X^s, D^s, R^s, A^s, G^s$

\For{$t \in 0\hdots|X^s|-1$}{ }
\State $q_t \gets Q([x_t, r_t, d_t], \theta)_{a_t}$
\State $q_{t+1}' \gets r_{t+t} + \beta_d d_{t+1} + \beta_g g_{t+1} + \max \gamma Q'([x_{t+1}, r_{t+1}, d_{t+1}], \theta')$
\EndFor
\State $\theta \gets \theta + \alpha\nabla \mathbb{E}\left[||q_a-q'||^2 \right] \quad\quad\text{Weight update}$
\EndFor
\State $\theta' \gets \theta \quad\quad\text{Target update}$
\EndFor
\EndProcedure
\end{algorithmic}
\end{algorithm}

\subsection{Details and Optional Enhancements}
For simplicity, several implementation details of the algorithm that do not change the underlying structure are not included in Algorithm \ref{algorithm}, but are described here instead. 
These are particular design choices that could be exchanged for alternative implementations. 

\paragraph{Recurrence}
Inputs to the Q function first pass through a recurrent network, which we chose to be an ESN for training efficiency and meta-learning performance based on our previous experiments.\cite{mckee2024reservoir}
The same considerations for handling recurrent states in offline RL must be made here as in other related work.\cite{kapturowski2018recurrent}
That is, during inference, continuity of recurrent states over episodes is maintained as the agent benefits from remembering its actions and results from previous episodes.
Because the ESN state is high dimensional, we recomputed states from observations during offline training, using the most recent hidden state from inference as the initial conditions.
By using the most recent hidden state, the offline training more directly synthesizes a plan for directing the agent from the real-time present context to its goal state.
%This goes in discussion
However, another strategy which may be more general over tasks and training schedules is to initialize the hidden state from zeros, then expend some number of observations in the replay buffer to ``warm up'' the hidden state until the effects of the initialization are negligible.

\paragraph{Density feedback embedding}
The agent is expected to learn nonlinear mappings of the normalized density value $\bar d_t$ to actions $a_t$.
To make that easier, a one-hot embedding represented binned subdomains of density over the unit interval was included in the feedback vector.
The number of bins was a hyperparameter.

%\paragraph{Soft memory storage thresholding}
%To limit the memory requirements of the model, a separate buffer was used to store only a representative sample of the observations for density calculation. This buffer was free to be a different size than the primary training buffer. A hyperparameter, the memory storage threshold, was included as the minimum negative density that must be exceeded for an observation to be included among memories used in density calculation. This ensured that no area of the observation space would be too redundantly represented in the sample and the number of samples would only increase with the progress of the explorer. An additional hyperparameter, the memory storage epsilon, adds back probabilistic storage for memories below that threshold, to avoid performance issues due to completely truncating the density function.

\paragraph{Prioritized Replay}
Episodes may be sampled either at random, by recency, or by performance.
A separate buffer, called the goal buffer, was included to contain only episodes in which the record for observation density was broken.
This way, a large number of exploration frontiers could be retained indefinitely and not forgotten due to prolonged periods without progress.
We added conditions so that the first minibatch of each training epoch was drawn from the goal buffer, the second from the end of the main buffer, and all others at random from the main buffer.

\paragraph{Goal states}
The term $g_{t+1}$ in the above code corresponds to density estimates with respect to all data collected so far.
To encourage faster expansion, the ``goal'' states, or minimum density states per episode, included a scalar multiplier on their density values.
Separate hyperparameters were used for the multipliers of episodes drawn from the main buffer and episodes drawn from the goal episode buffer.
This method incentivizes exploration of both local and global memory density minima.

\paragraph{Memory clustering for density estimation}
To estimate the density of new observations, a buffer of representative task states was required.
To keep the buffer small, we implement the RECODE clustering algorithm\cite{saade2023unlocking}.
This algorithm dynamically chooses whether to form a new cluster centroid from the current observation or to merge that observation with the nearest existing centroid.
This allows the agent to distribute a minimal number of cluster centroids along its exploration paths, keeping estimation cheap and practical for high dimensional task spaces.

\section*{Results}

\begin{figure}[!p]
    \begin{subfigure}[t]{\textwidth}
        \centering
        \includegraphics[width=.725\linewidth]{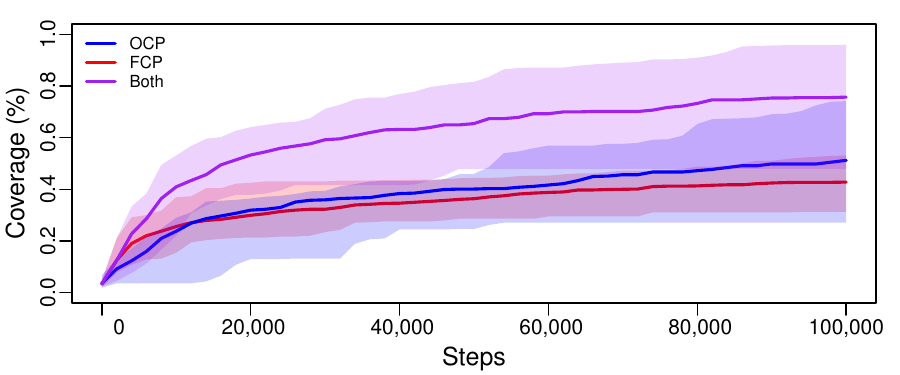}
        \caption{Coverage curves for fixed maze}
        \label{fig:results1}
    \end{subfigure}
    
    \vspace{0.4cm}
    \begin{subfigure}[t]{\textwidth}
        \centering
        \includegraphics[width=.725\linewidth]{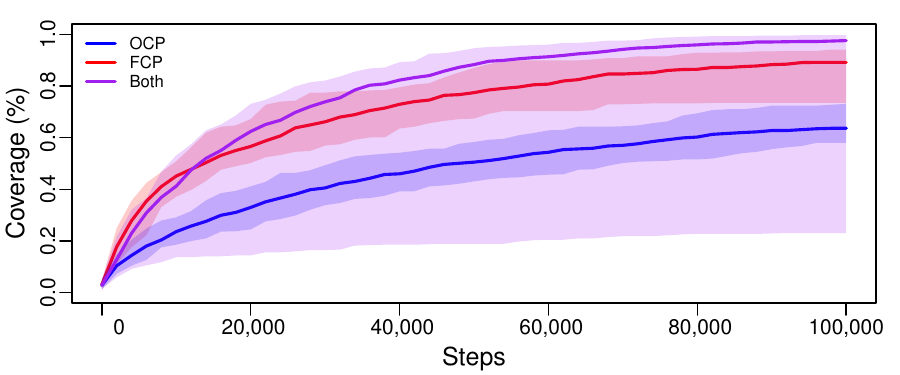}
        \caption{Coverage curves for random maze}
        \label{fig:results2}
    \end{subfigure}
    \vspace{0.4cm}
    
    \begin{subfigure}[t]{\textwidth}
        \centering
        \includegraphics[width=.725\linewidth]{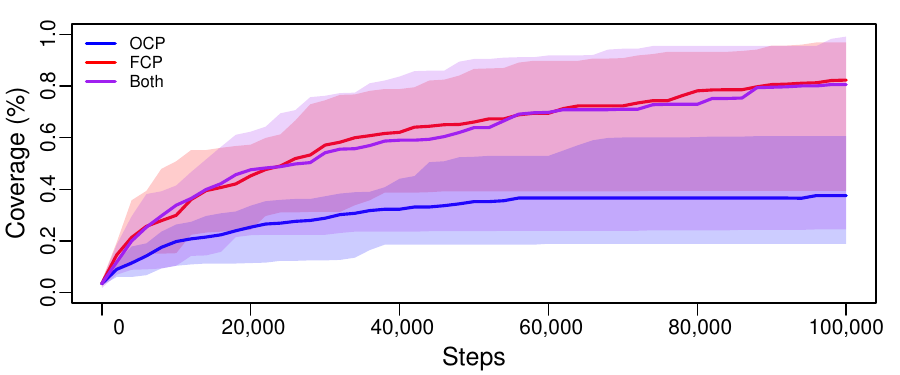}
        \caption{Coverage curves for continual maze}
        \label{fig:results3}
    \end{subfigure}
    \begin{subfigure}[t]{\textwidth}
        \centering
        \includegraphics[width=.725\linewidth]{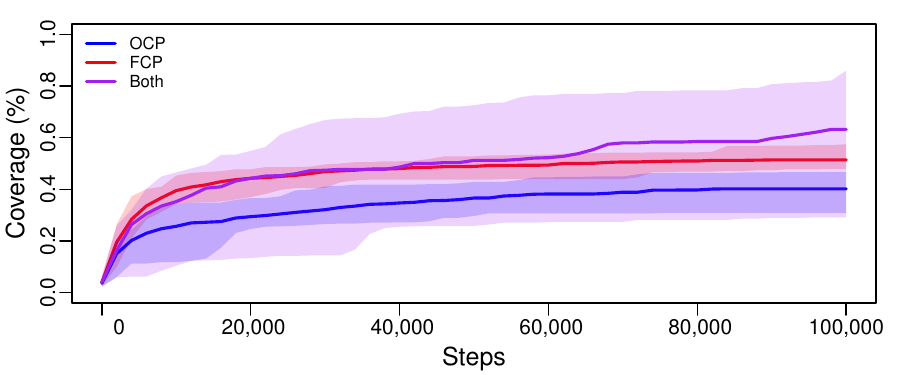}
        \caption{Coverage curves for noisy maze}
        \label{fig:results4}
    \end{subfigure}

    \label{fig:results_curves}
    \caption{Coverage curves for all three tasks. Line shows median trajectory and shaded regions span minimum to maximum scores.}
\end{figure}

\begin{figure}[!p]
    \centering
    \begin{subfigure}[t]{0.275\textwidth}
        \includegraphics[width=\linewidth]{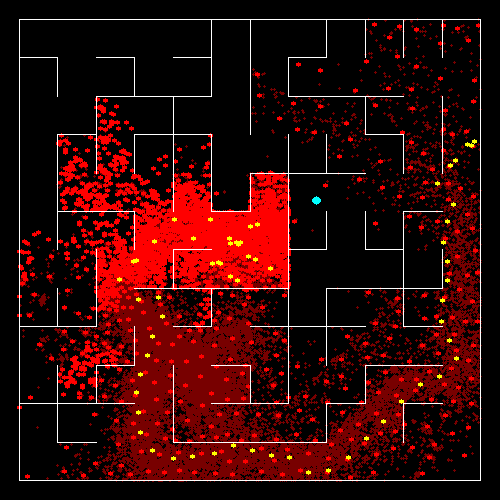}
        \caption{OCP only}
        \label{fig:results1ocp}
    \end{subfigure}
    \begin{subfigure}[t]{0.275\textwidth}
        \includegraphics[width=\linewidth]{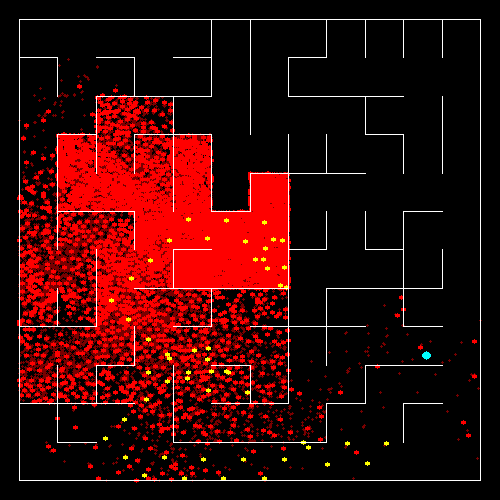}
        \caption{FCP only}
        \label{fig:results1fcp}
    \end{subfigure}
    \begin{subfigure}[t]{0.275\textwidth}
        \includegraphics[width=\linewidth]{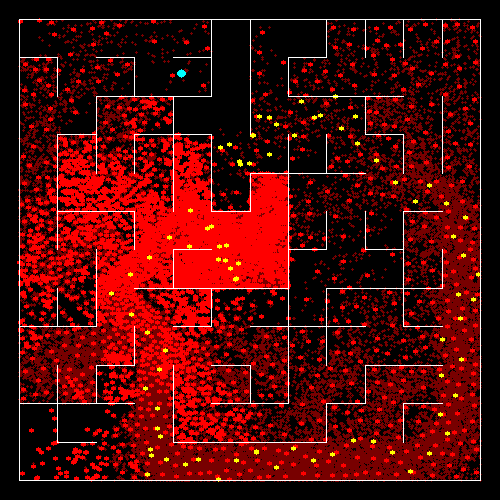}
        \caption{Combined model}
        \label{fig:results1both}
    \end{subfigure}
    \caption{Best example results for a fixed maze.}
    \label{fig:results1_pics}
    
    \vspace{0.5cm}
    \begin{subfigure}[t]{0.275\textwidth}
    \includegraphics[width=\linewidth]{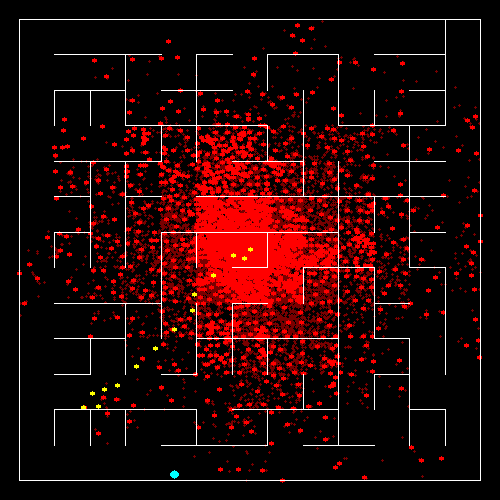}
    \caption{OCP only}
        \label{fig:results2ocp}
    \end{subfigure}
    \begin{subfigure}[t]{0.275\textwidth}
        \includegraphics[width=\linewidth]{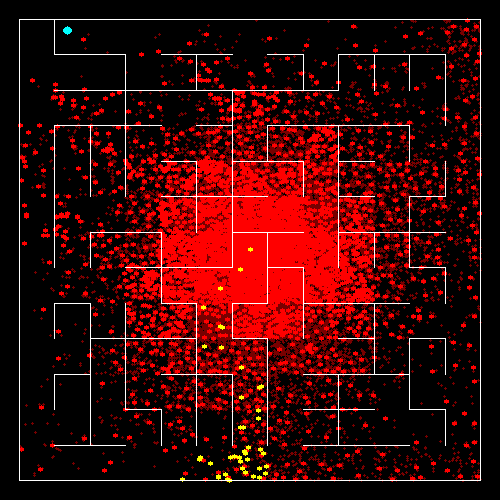}
        \caption{FCP only}
        \label{fig:results2fcp}
    \end{subfigure}
    \begin{subfigure}[t]{0.275\textwidth}
        \includegraphics[width=\linewidth]{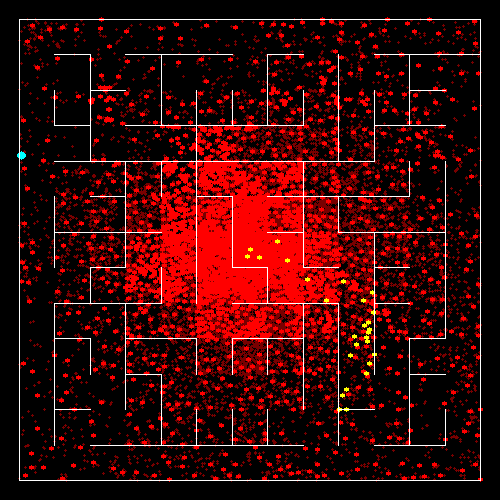}
        \caption{Combined model}
        \label{fig:results2both}
    \end{subfigure}
    \caption{Best example results for randomized mazes.}
    \label{fig:results2_pics}

    \centering
    \begin{subfigure}[t]{0.275\textwidth}
    \includegraphics[width=\linewidth]{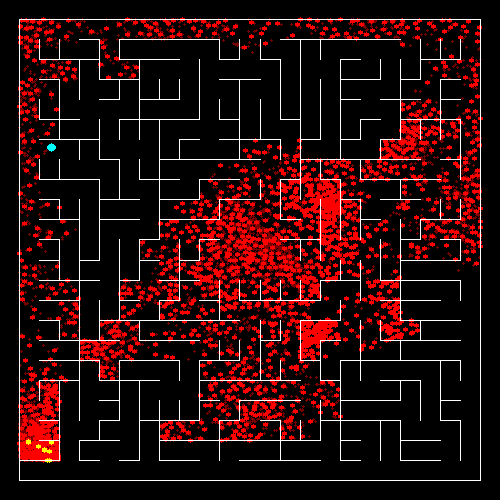}
    \caption{OCP only}
        \label{fig:results3ocp}
    \end{subfigure}
    \begin{subfigure}[t]{0.275\textwidth}
        \includegraphics[width=\linewidth]{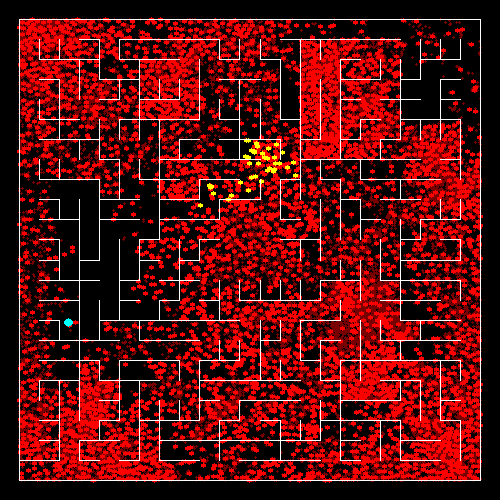}
        \caption{FCP only}
        \label{fig:results3fcp}
    \end{subfigure}
    \begin{subfigure}[t]{0.275\textwidth}
        \includegraphics[width=\linewidth]{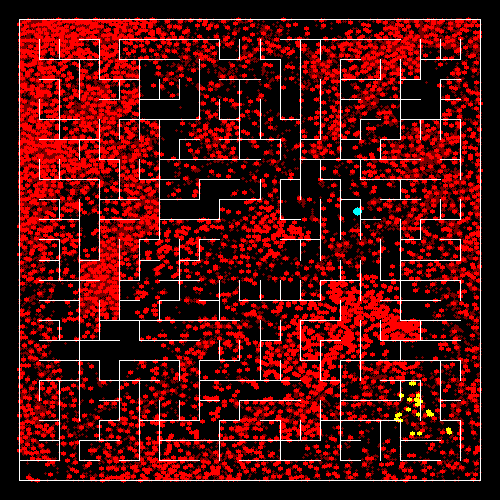}
        \caption{Combined model}
        \label{fig:results3both}
    \end{subfigure}
    \caption{Best example results for a fixed continual maze.}
    \label{fig:results3_pics}

    \centering
    \begin{subfigure}[t]{0.275\textwidth}
    \includegraphics[width=\linewidth]{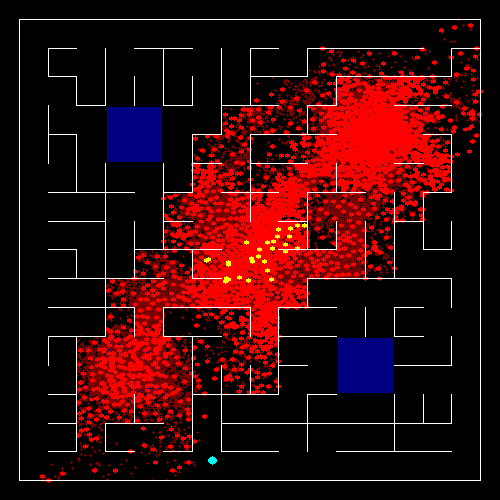}
    \caption{OCP only}
        \label{fig:results4ocp}
    \end{subfigure}
    \begin{subfigure}[t]{0.275\textwidth}
        \includegraphics[width=\linewidth]{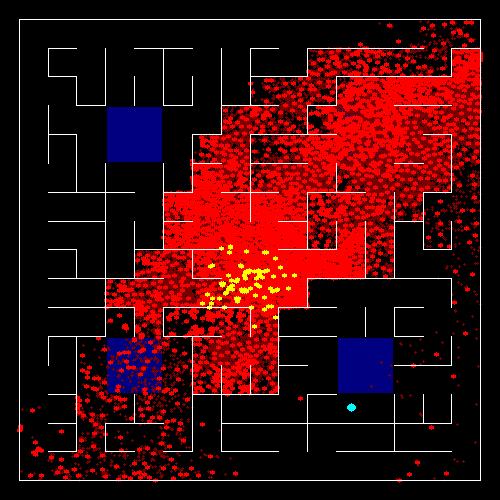}
        \caption{FCP only}
        \label{fig:results4fcp}
    \end{subfigure}
    \begin{subfigure}[t]{0.275\textwidth}
        \includegraphics[width=\linewidth]{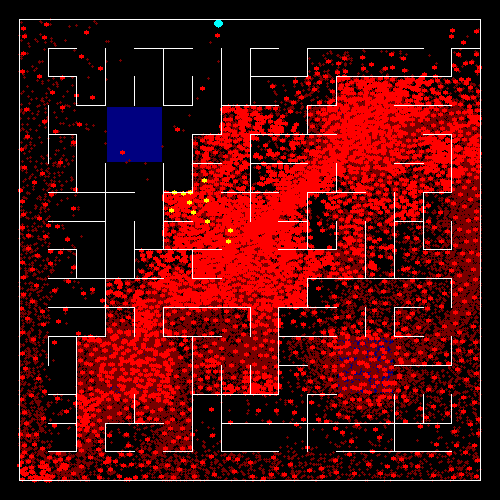}
        \caption{Combined model}
        \label{fig:results4both}
    \end{subfigure}
    \caption{Best example results for a maze with noise traps.}
    \label{fig:results4_pics}
\end{figure}

The results for each experiment are presented below. 
Each training curve is shows the media trajectory over 32 runs per condition.
The groups were the continuous (DDPG) version of the model (1) observation-conditioned policy (OCP), (2) feedback-conditioned policy (FCP), and (3) a policy that conditions on both observations and feedback.
By comparing these models we can examine the overall benefit to exploration added by the meta-learning aspect (i.e., FCP) over original Go-Explore type algorithms (OCP).
The best example runs from each group are also shown for each experiment, though the worst performing groups in some cases included examples that made almost no progress.

%Fixed maze
\paragraph{Fixed maze}
The fixed maze results compared are shown in Figure \ref{fig:results1}.
The top coverage for each group was 74\% for OCP, 53\% FCP, and 96\% for combined.
The bottom scores excluding pathological cases (<15\%) were 27\% for OCP, 31\% for FCP, and 48\% for combined. 
This demonstrates that while observations were important for exploring the fixed maze, where everything served as a consistent landmark, internal feedback produced much more efficient exploration.

The best runs from each group are shown in Figure \ref{fig:results1_pics}.
Large red dots represent the centroids created by the RECODE algorithm and used in density estimation.
Small dark red dots show the actual locations visited by the agent over the course of its lifetime.
The yellow dots shows its locations from only the final episode.
The larger cyan dot shows the lowest density point that the agent visited overall, computed at the end of the experiment.

\paragraph{Random maze}
Results for the randomized mazes are shown in Figure \ref{fig:results2}.
Here the combined model performed best but only by a small margin from the FCP.
The OCP performed significantly worse than either. 
These results demonstrate that when paths to the frontier are inconsisent, observations cannot be relied upon to determine the best actions.
The top scores were 73\% for OCP, 94\% for FCP, and 100\% for the combined model.
The bottom scores were 58\% for OCP, 73\% for FCP, and 23\% for the combined model

Images from the top performers in each group are shown in Figure \ref{fig:results2_pics}.
It is clear that the OCP was the most densely distributed around its starting point, with very few points covering the outer edges of the maze. 
Some irregular concentrations of points are also apparent.
The FCP model covered almost the entire maze much more evenly.
The combined model produced the most uniform maze coverage, showing clear exploitation of the maze perimeter, which was only blocked at one corner.

\paragraph{Continual Maze}
Results for the continual maze are shown in Figure \ref{fig:results3}.
Like the random mazes, the FCP and combined models greatly outperformed the OCP model, but while performing nearly equally to each other.
The top scores were 61\% for OCP, 97\% for FCP, and 99\% for the combined model.
The bottom scores were 19\% for the OCP, 39\% for FCP, and 24\% for the combined model.

Figure \ref{fig:results3_pics} shows the top performers in each group. 
The OCP model explores widely but demonstrates many areas of high density, indicating that it frequently became stuck.
The FCP and combined models show much more efficient exploration and coverage overall.

The continual maze was scaled to the same overall size as the fixed maze, resulting in much narrower corridors, so these results were obtained using a smaller RECODE parameter $\kappa=0.025$ as opposed to 0.2, allowing cluster centroids to be closer together.

\paragraph{Noisy Maze}
Results for the noisy maze are shown in Figure \ref{fig:results4}.
All three models performed significantly worse as a result of the noise traps.
The OCP performed worst, FCP slightly better, and combined model best.
The top scores were 47\% for OCP, 57\% for FCP, and 86\% for the combined model.
The bottom scores were 31\% for OCP, 48\% for FCP, and 29\% for the combined model.

\paragraph{Non-Euclidean Graph Maze}

\begin{figure}[!h]
    \centering
    \includegraphics[width=.725\linewidth]{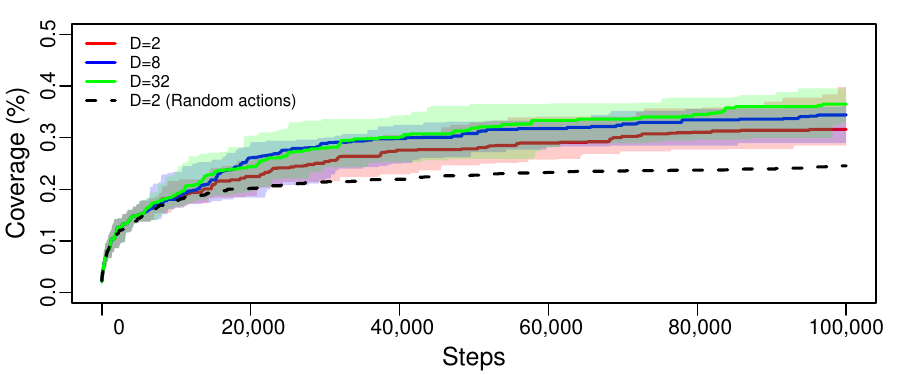}
    \caption{Coverage curves for non-Euclidean graph mazes of varying dimension.}
    \label{fig:results_graphenv}
\end{figure}
    
Results on the non-Euclidean graph mazes of dimension 2, 8 and 32 show minor differences in coverage with overlapping variances (Figure \ref{fig:results_graphenv})
All runs outperformed a random explorer.
Higher dimensional mazes tended to have higher coverage, contradicting concerns about the usefulness of the distance metric as dimensionality increases \cite{aggarwal2001surprising}.
Potentially these differences are more reflective of differences in the resulting reward function than problems with the distance metric per se.

\paragraph{Overall trends}
Some overall patterns are apparent from these results.
First, that in all but the fixed maze, the FCP outperforms the OCP, suggesting that the feedback-based tactics employed by the agent for exploring are more important than observable references, particularly when those references are unreliable due to randomization, complexity, or noise.
Second, conditioning the policy on both observations and feedback is the most general model, as tasks in general will have a mix of reliable and unreliable states.
Third, the combined model in some cases had greater variance than OCP or FCP, possibly resulting the greater complexity involved in balancing the two objectives.

\paragraph{Generalization experiments}
\begin{figure}[!h]
    \begin{subfigure}[t]{\textwidth}
        \centering
        \includegraphics[width=.725\linewidth]{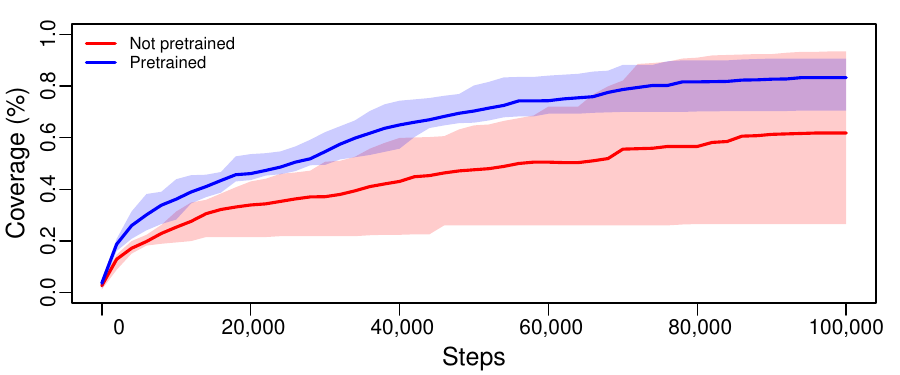}
        \caption{Generalization results for a small maze ($13\times13$).}
        \label{fig:results_gen_sm}
    \end{subfigure}

    \begin{subfigure}[t]{\textwidth}
        \centering
        \includegraphics[width=.725\linewidth]{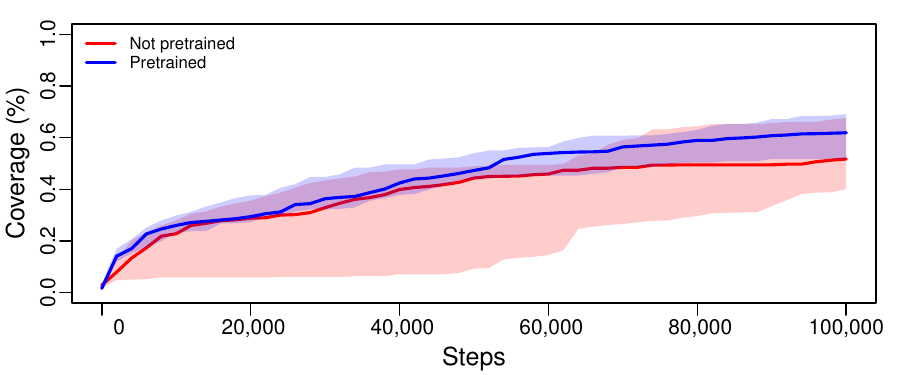}
        \caption{Generalization results for a large maze ($17\times17$).}
        \label{fig:results_gen_lg}
    \end{subfigure}
    \caption{Results for generalization test comparing a pretrained model to an untrained model.}
    \label{fig:results_gen}
\end{figure}

To test for the presence of generalizable exploration skills in the trained model, we deployed the best performing model from the fixed-maze experiment and an untrained model for comparison.
The first maze was about the same size as the training maze ($12\times12$), and the second was much larger ($17\times17$).
Results are shown in Figure \ref{fig:results_gen}.
The pretrained maze achieved higher coverage on average with much greater reliability.
The untrained model occasionally matched the high score of the pretrained model in both tests.

\section{Discussion}

%Summary
%--Overview
Our experiments demonstrate a clear benefit to training agents to both plan a path to their exploration frontier and to optimize their coverage of novel task states in real-time.
The latter training was achieved by feeding the agent online density calculations corresponding to the current observation while also training it to minimize those density estimates.
The resulting agent both learns to use observable landmarks to determine its exploration frontier, while exploring up to and beyond that frontier by observing the relative changes in density resulting from each action.
We observed that the process learned by the agent is akin to gradient descent over memory density with respect to its actions, a complementary strategy to goal-driven exploration.
Importantly, the added functionality allows the agent to perform in a much wider range of conditions than afforded by the return-then-explore concept on its own.
Our agent explores fixed environments, highly random and unpredictable environments, continual environments that do not reset, and environments with intrinsic noise.
Each of these challenges requires unique navigational skills that are promoted by the combined objective functions and internally generated feedback.

%--Limitations
\paragraph{Combining exploration and intrinsic reward}
In principle, RL-based exploration algorithms can incorporate extrinsic reward functions to accelerate training on tasks with sparse rewards.
In practice, there are challenges that we have left out of the scope of this paper, namely conflict between the best hyperparameter values for complex tasks with sparse rewards and those best for exploration.
Here we found that a higher learning rate ($3e-4$), polyak coefficient on target updates ($1.0$, or $100\%$), and frequency of target updates (every optimization step) are not those generally recommended by the authors of DQN or most other applications \cite{Mnih2015}.
The next steps for this work are to investigate the best ways to integrate the exploration algorithm into other tasks.

\paragraph{Transforming observations and data compression}
The current agent computes density $\mathcal{D}(x)$ for its reward function. 
The general case is to consider $\mathcal{D}(f(x))$, where $f$ is some transformation of the data.
At a minimum, some elements of the observation vector may be unimportant and can be shrunken, reducing their effect on the agent’s exploration path.
Others may be more important and in need of up-scaling to reach appropriate influence on the agent. 
In more complex cases, particular vectors or extracted features may be important, like the number of a key object, a score, or small indicators as with most heads-up displays. 
But we cannot say what is important in general, so it is not clear what kind of function $f$ should be or what it should be trained to optimize across all use cases.
One possible heuristic is to batch normalize all elements to standard normal distributions such that everything is given uniform weight in exploration.
But this is not satisfying where there are a very large number of elements, most of which are likely mundane.
For instance, many games involve a small foreground consisting of characters or symbols, and a much larger background consisting of repeated tiles or textures, such as in the benchmarking game Crafter.\cite{hafner2021benchmarking}

Other exploration algorithms, Go-Explore\cite{ecoffet2019go} and Latent Go-Explore\cite{gallouedec2023cell}, use different methods of clustering and quantized autoencoding\cite{van2017neural} to summarize general domains of the observations, which can then be used to determine novelty and guide exploration. 
In preliminary results on the same maze environments tested here, we have found that the clustering algorithm RECODE\cite{saade2023unlocking} maintained the algorithm's exploration performance while representing 100,000 observations in just 6000 centroids, making it practical for game environments with pixel input and other high dimensional tasks.
But there is no general principle for choosing a data reduction method, and task-specific structures must be taken into consideration.
For instance, a common critique of the original Go-Explore implementations (and hence related cluster-based implementations) is that it was designed primarily to set records on Montezuma's Revenge, which involves discrete scenes with large pixel variation between, but only small pixel variation within them.
Latent Go-Explore \cite{gallouedec2023cell} aims for a more general solution with quantized auto-encoding.
However, if progress on a task is represented by samples from a unimodal distribution, then clustering algorithms may have little benefit or even hurt performance by under-representing important differences between observations.

\paragraph{Conclusions}
In this study, we demonstrated a major improvement to RL-based exploration.
We find that the return-then-explore concept is inadequate in (1) randomized environments for which return paths are unpredictable and (2) continual environments for which returning is a problem of inverting previous navigation, rather than simply resetting to a start point.
We introduced a meta-learning approach in which the agent uses real-time feedback on the novelty of its observations to continue maximizing its progress, even after it has left familiar regions of the task space. 
This method leads to major improvements on the original fixed maze used to test return-then-explore methods, on randomized mazes, mazes without episodic resetting of position, and mazes with noisy regions.
This approach is thus suitable for efficiently exploring task spaces in general.

\bibliographystyle{unsrtnat}
\bibliography{refs}

\newpage
\section{Appendix A: Continuous Action Space}
The training algorithm for continuous action spaces for discrete spaces was DQN and for continuous action spaces was DDPG.
In principle, other training algorithms should work, but offline Q-learning is chosen here because the policies are expected to be nonstationary, and so they benefit from both cumulative offline training and simplicity of the algorithm.
For pure exploration, the task reward is swapped entirely with the normalized negative density values.
If augmenting an RL task, the final reward function for training is a weighted combination of negative density and the original task reward.   
\begin{align}
z^F_t &= \text{ESN}([a_{t-1},r_{t-1}, D(x_t)]) \\
z^X_t &= \text{ESN}(x_t) \\
q^F_t &= \text{MLP}(z^F_t) \\
q^X_t &= \text{MLP}(z^X_t) \\
q^{F\prime}_t &= r_t + \gamma q^F_t(z_t') \\
q^{X\prime}_t &= r_t + \gamma q^X_t(z_t') 
\end{align}
\begin{figure}[!ht]
    \centering
    \includegraphics[width=\linewidth]{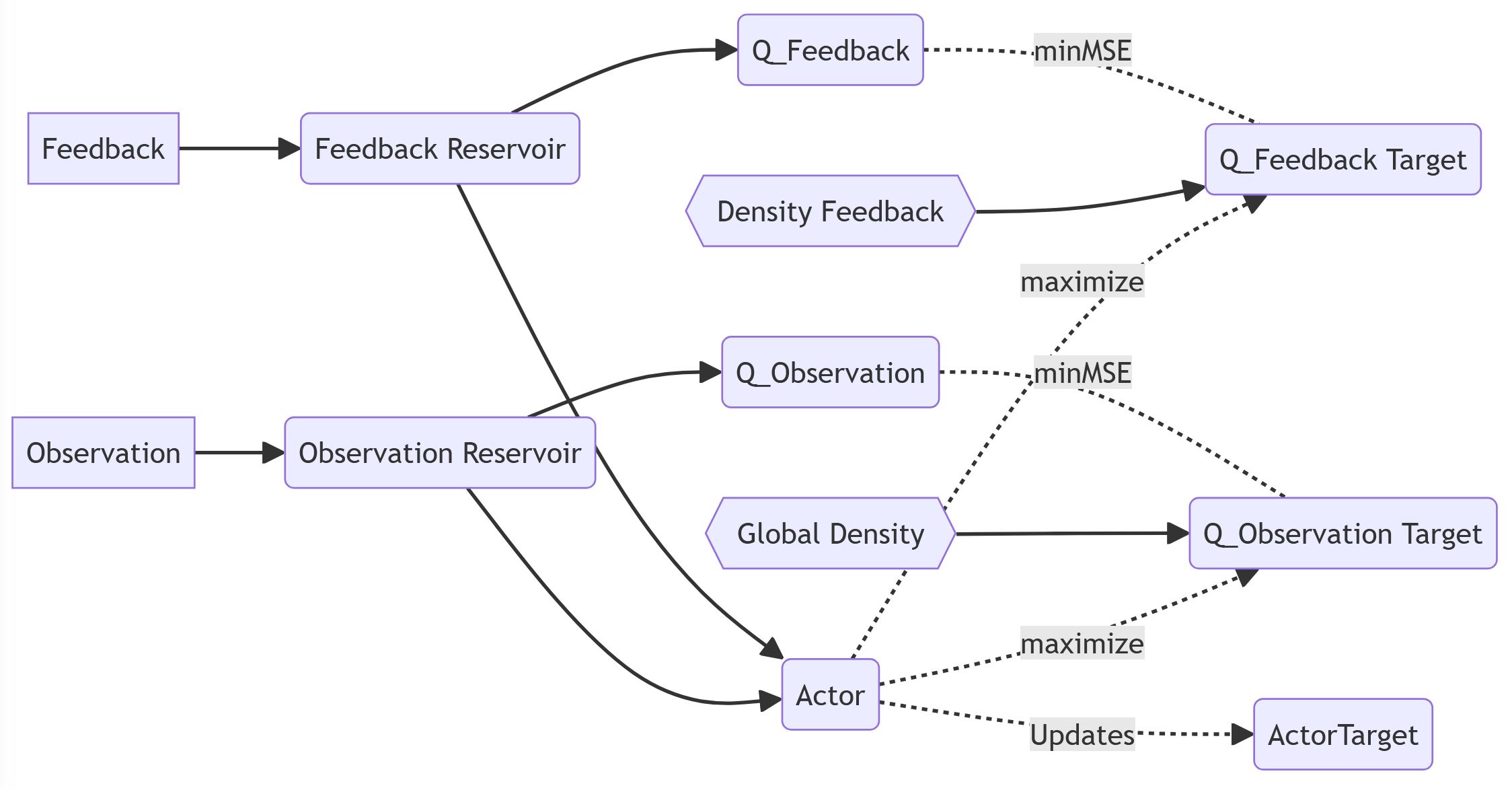}
    \caption{DDPG based explorer for continuous actions with separated feedback and observation reservoirs and Q functions.}
    \label{fig:model_ddpg}
\end{figure}
When adapting the algorithm with DDPG, the model was set up with two independent reservoirs and Q networks corresponding to feedback and observations.
This was done because the feedback objective was computed online, while the global goal objective was computed offline.
The two halves of the model were expected to have different convergence properties, with the feedback network converging to stationary policy and the observation network continually drifting to incorporate new features corresponding to the exploration frontier.

\newpage
\section{Appendix B: DDPG explorer hyperparameters}
The following parameters were used to obtain the experimental results with the DDPG based explorer on the continuous maze tasks:

\begin{table}[!h]
    \centering
    \begin{tabular}{r|l}
    \hline
    Hyperparameter & Value \\ 
    \hline
    \multicolumn{2}{l}{\textbf{Objectives and gradients}} \\
    \hline
    Learning rate & 3e-4 \\
    Discount $\gamma$ & 0.9 \\
    Episode goal reward scale $\beta_d$ & 5.0 \\
    Global goal reward scale $\beta_g$ & 5.0 \\
    Training steps per epoch & 100 \\
    Target updates per training epoch & 10 \\
    Target update polyak coefficient & 1.0 (100\% updated)\\
    Optimizer & RMSProp \\
    \hline
    \multicolumn{2}{l}{\textbf{Training}} \\
    \hline
    Replay buffer size & 20,000 \\
    Goal buffer size & 20,000 \\
    Minibatch size & 200 \\
    \hline
    \multicolumn{2}{l}{\textbf{Model}} \\
    \hline
    MLP Hidden layer size & 256 \\
    MLP Hidden layers & 3 \\
    Observation ESN size & 60*Obs dimension \\
    Feedback ESN size & 180 \\

    ESN spectral radius & 1.15 \\
    Density \textit{k} & 15 \\  
    Density feedback bins &  8\\
    \hline
    \multicolumn{2}{l}{\textbf{Epsilon-greedy policy}} \\
    \hline
    Epsilon initial probability & 1.0 \\
    Epsilon minimum probability & 0.1 \\
    Epsilon decay constant & 0.9 \\
    Initial steps with epsilon = 1.0 & 100 \\
    \hline
    \multicolumn{2}{l}{\textbf{RECODE}} \\
    \hline
    $\gamma$ & 1.0 \\
    $\kappa$ & 0.2 \\
    Insertion probability & 0.1 \\
    Decay & 0.9999 \\
    Buffer size & 6000 \\
    \hline
    \end{tabular}
    \caption{Hyperparameters for continuous (DDPG) based explorer algorithm used in the current experiments.}
    \label{tab:hyperpars}
\end{table}

\end{document}